\documentclass[sigconf]{acmart}
\usepackage{graphicx}
\usepackage{comment}

\usepackage{amsmath,amssymb} 
\usepackage{color}
\usepackage{algpseudocode}
\usepackage{subfigure}
\usepackage{multirow} 
\usepackage{balance} 
\usepackage[linesnumbered,ruled]{algorithm2e}

\AtBeginDocument{%
  \providecommand\BibTeX{{%
    \normalfont B\kern-0.5em{\scshape i\kern-0.25em b}\kern-0.8em\TeX}}}

\copyrightyear{2021}
\acmYear{2021}
\setcopyright{acmcopyright}
\acmConference[MM '21]{Proceedings of the 29th ACM International Conference on Multimedia}{October 20--24, 2021}{Virtual Event, China}
\acmBooktitle{Proceedings of the 29th ACM International Conference on Multimedia (MM '21), October 20--24, 2021, Virtual Event, China}
\acmPrice{15.00}
\acmDOI{10.1145/3474085.3475179}
\acmISBN{978-1-4503-8651-7/21/10}

\begin{document}
\fancyhead{}

\title{Semi-Autoregressive Image Captioning}


\author{Xu Yan$^{1,2,\dagger}$, Zhengcong Fei$^{1,2,\dagger}$, Zekang Li$^{1,2}$, Shuhui Wang$^{1, *}$, Qingming Huang$^{1,2,3}$, Qi Tian$^{4}$}
\thanks{$^\dagger$These authors contribute equally to this work.  $^\ast$Corresponding author.}
\affiliation{%
	\institution{$^1$ Key Lab of Intell. Info. Process., Inst. of Comput. Tech., CAS, Beijing, China}
	\institution{$^2$ University of Chinese Academy of Sciences, Beijing, China}
	\institution{$^3$ Peng Cheng Laboratory, Shenzhen, China}
	\institution{$^4$ Cloud BU,  Huawei Technologies, Shenzhen, China}
}
\email{{yanxu19s, feizhengcong, lizekang19g,  wangshuhui}@ict.ac.cn, qmhuang@ucas.ac.cn, tian.qi1@huawei.com}

\begin{abstract}
	Current state-of-the-art approaches for image captioning typically adopt an autoregressive manner, \emph{i.e.}, generating descriptions word by word, which suffers from slow decoding issue and becomes a bottleneck in real-time applications. Non-autoregressive image captioning with continuous iterative refinement, which eliminates the sequential dependence in a sentence generation, can achieve comparable performance to the autoregressive counterparts with a considerable acceleration. Nevertheless, based on a well-designed experiment, we empirically proved that iteration times can be effectively reduced when providing sufficient prior knowledge for the language decoder. Towards that end, we propose a novel two-stage framework, referred to as Semi-Autoregressive Image Captioning (SAIC), to make a better trade-off between performance and speed. The proposed SAIC model maintains autoregressive property in global but relieves it in local. Specifically, SAIC model first jumpily generates an intermittent sequence in an autoregressive manner, that is, it predicts the first word in every word group in order. Then, with the help of the partially deterministic prior information and image features, SAIC model non-autoregressively fills all the skipped words with one iteration.  Experimental results on the MS COCO benchmark demonstrate that our SAIC model outperforms the preceding non-autoregressive image captioning models while obtaining a competitive inference speedup. Code is available at \href{https://github.com/feizc/SAIC}{https://github.com/feizc/SAIC}.
\end{abstract}



\maketitle

\section{Introduction}

Image captioning is one of the fundamental tasks in multimedia analysis and computer vision, which aims to generate a natural description for the visual content of the given image automatically. Most image captioning systems follow an encoder-decoder paradigm \cite{sutskever2014sequence,Vinyals2015Show,Xu2015Show,chen2015mind,Anderson2017Bottom,exploring,huang2019attention,cornia2020meshed}. Among these methods, the 
visual encoder, \emph{e.g.}, convolutional neural network (CNN), first extracts features from the input image. The descriptive sentence is then decoded according to these refined features, one word at each time, using a recurrent neural network (RNN). 
Despite their remarkable performance, the sequential nature of text generation, \emph{i.e.}, word by word, makes the decoding procedure not parallelizable and results in a high latency, which is challenging to work effectively in some real-time production applications \cite{guo2020non}.
Inspired from neural machine translation \cite{nat}, one straight forward solution is the non-autoregressive image captioning (NAIC) \cite{nat,fei2019fast}, which predicts the entire sentence in one shot. However, such a one-pass NAIC model usually lacks dependencies between words and struggles to produce smooth and accurate descriptions.

\begin{figure}
	\begin{center}
		\includegraphics[width=1\columnwidth]{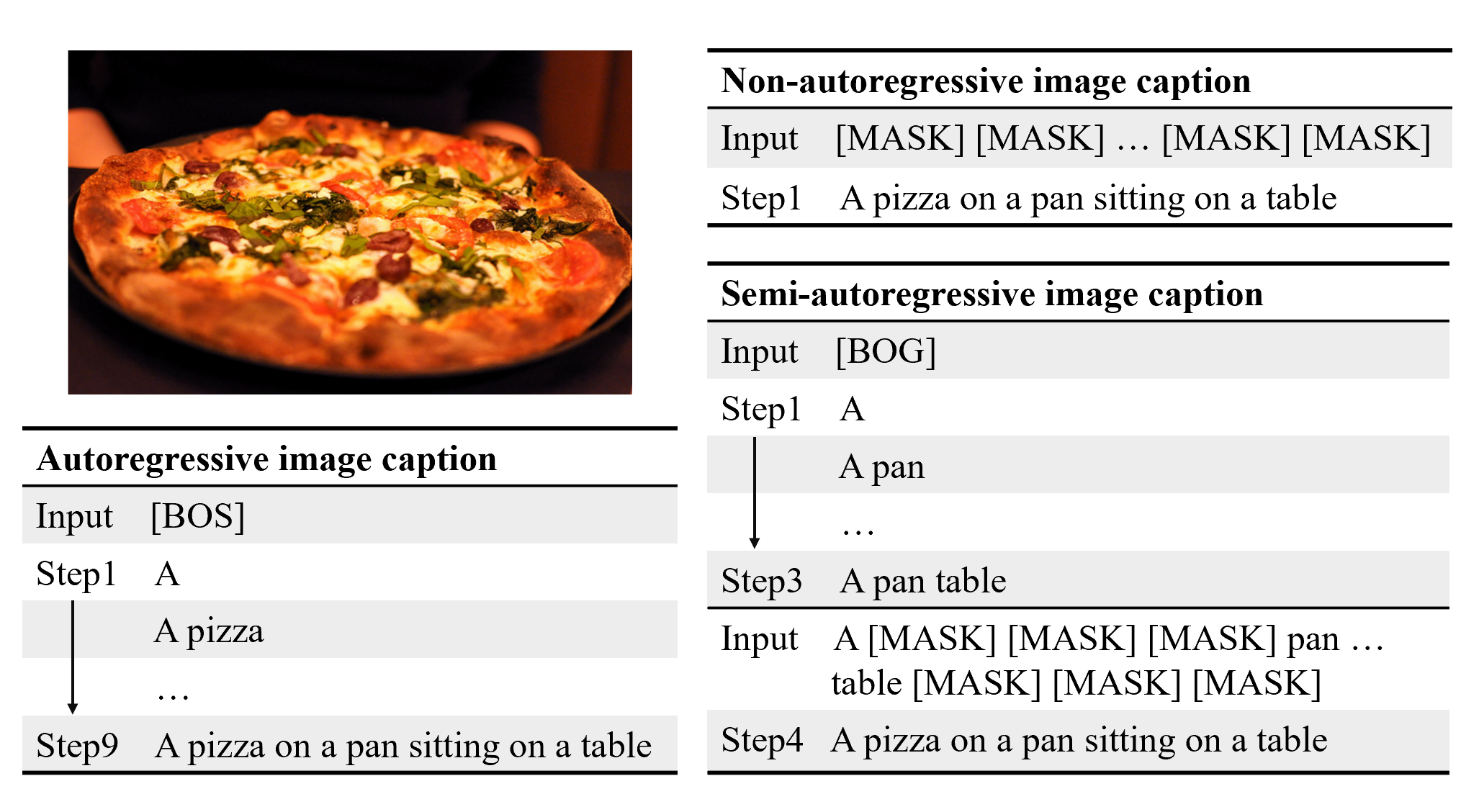}
	\end{center}
	\caption{Illustration of AIC, NAIC and SAIC. AIC models generate the next word conditioned on all preceding subsentence, while NAIC models output all words in parallel in one step. Comparatively, our SAIC model considers a caption as a sequence of word groups. The first word of each group is generated one-by-one while the remaining words are predicted simultaneously conditioned on both the visual representation and jumpily generated words.}
	\label{fig:1}
\end{figure}

Recent studies show that extending one-pass NAIC to constant multi-pass, called iterative refinement (IR-NAIC), is promising to break the captioning performance dilemma \cite{dir,gao2019masked,fei2020iterative,yang2019non}. Unlike one-pass NAIC, which outputs the total descriptions immediately, IR-NAIC takes the caption hypothesis from the previous iteration as a reference and regularly modifies the new sentence, with a masking strategy or fusion network. The refinement process is terminated until reaching a pre-defined iteration count or no changes appear in the new sentence. Compared with conventional autoregressive image captioning (AIC), IR-NAIC with three iteration steps runs an average $\times$3 times faster with a comparable captioning quality, as reported by \cite{fei2020iterative}.

However, in this paper, we highlight that decoding with multiple iterative refinements in NAIC is unnecessary when providing a good partial prior knowledge. To verify this statement, we carefully design an experiment to understand the relationship between segmental dependency context and iteration times. In practice, we first mask some proportion of words on the caption hypothesis generated by a trained AIC model with different strategies, \emph{i.e.}, \emph{Head, Tail, Random}, and \emph{Group}. Then the residual segments are taken as the input to the language decoder, aiming to measure the quality of the final output caption. Surprisingly, we observed that even masking 70\% of AIC hypothesis, the remaining prior knowledge can still help one-shot NAIC model compete with the standard IR-NAIC model. This result confirms a promising improvement direction about the form of iteration.

Inspired by this, we propose a novel two-stage framework, referred to as Semi-Autoregressive Image Captioning (SAIC), which combines the advantage of both AIC and NAIC. After extracting the visual representations with the visual encoder, SAIC first utilizes an autoregressive decoder, named Outliner, to produce several partial discontinuous words in a caption. Then a non-autoregressive decoder, named Filler, predicts previously skipped words with one iteration according to the deterministic prior knowledge. Since both the Outliner and Filler share the same model architecture and parameters, SAIC does not increase extra parameters significantly. To train the SAIC model effectively and efficiently, we further propose three training techniques including group-aware sampling, curriculum learning \cite{bengio2009curriculum} as well as hybrid knowledge distillation. Experiment results on the MS COCO dataset show that our proposed SAIC brings a consistent decoding speedup relative to the autoregressive counterpart while gains far superior performance than the state-of-the-art IR-NAIC models. 

To sum up, our main contributions are as follows: 
\begin{itemize}
	\item Through a well-designed experiment, we demonstrate that the number of IR-NAIC can be significantly reduced when providing a good prior knowledge.
	\item We propose a new two-stage semi-autoregressive framework to take a better trade-off between caption generation speed and quality performance. To be specific, Outliner first jumpily generates a series of discontinuous words in an autoregressive manner, and Filler then pads all previously skipped words in one non-autoregressive step.
	\item We introduce three strategies to improve the model training procedure: group-aware sampling, curriculum learning, and hybrid knowledge distillation.  Experimentally, SAIC is able to decode faster than the sequential counterpart while strikingly narrowing the performance gap.
\end{itemize}

\section{Background}

\subsection{Autoregressive Image Caption} 

Autoregressive generation is the mainstream approach in conventional image captioning, which decomposes the distribution of target sentence $P(S|I)$ into a chain of conditional probabilities in a directional manner as:
\begin{equation}
	P(S|I) = \prod_{t=1}^{N}P(w_t|I, w_{<t}), \label{eq:1}
\end{equation}
where $w_{<t}$ denotes the generated historical sub-sentence before time step $t$. In particular, beam search  \cite{wiseman2016sequence,vijayakumar2016diverse} is commonly used as a heuristic search technique, because it maintains multiple hypotheses at each decoding step and leads to a satisfactory captioning performance. 
However, the existence of condition $w_{<t}$ requires that the AIC model have to wait for $w_{t-1}$ to be produced before predicting current $w_t$, which hinders the possibility of parallel computation along with time step.

\subsection{Non-autoregressive Image Caption} 

Regardless of its effectiveness, the sequential inference methods have two major drawbacks. One is that it cannot generate multiple words simultaneously, leading to inefficient use of parallel computing hardware such as GPUs \cite{ren2020study}. The other is that beam search has been found to output low-quality when applied to large search spaces \cite{vijayakumar2016diverse}.
Non-autoregressive method is first proposed by \cite{nat,gao2019masked} to address the above issues, allowing the image captioning model to generate all target words simultaneously. NAIC replaces $w_{<t}$ with independent latent variable $z$ to remove the sequential dependencies and rewrite Equation \ref{eq:1} as:
\begin{equation}
	P(S|I) = P(N|I) \prod_{t=1}^{N}P(w_t|I, z).
\end{equation}
Since words are generated independently from each other during the entire decoding process, it usually results in duplicated or missing words in the obtained sentences.
Subsequently, researchers developed more advanced methods to enhance the modeling of $z$, such as reordered latent variable \cite{fei2019fast} and objective function optimization \cite{guo2020non}, but there still exists a significant performance gap between AIC and NAIC.

\subsection{Iterative Refinement based Non-autoregressive Image Caption}

The previous one-pass NAIC can be further boosted by introducing the multi-pass refinement mechanism \cite{gao2019masked,fei2020iterative,gao2019deliberate}. Specifically, IR-NAIC applies a fusion function $f$ to deal with the sentence $S'$ produced in the preceding stage and comprehensively predict the new sentence $S$ by:
\begin{equation}
	P(S|I) = \prod_{t=1}^{N'} P(w_{m(t)}|I,f(S')),
\end{equation}
where $N'$ corresponds to the number of newly generated words in $S'$, and $m(t)$ is the real position of $t$-th refined word in $S$. In this way, the sentence generation process of IR-NAIC can be concluded as: first, the NAIC model produces a coarse caption as the initial hypothesis and then iteratively refines it until reaches a pre-set criteria with constant steps. 
In this work, we utilize masked prediction \cite{gao2019masked,dir} as the representation of IR-NAIC due to its excellent performance and simplicity, where tokens are randomly masked in training but selected with low confidences during inference. 


\section{Is Iterative Refinement All You Need?}

Previous work has pointed out that heuristic NAIC with multiple iterations can accelerate the overall speed to a significant extent, however, it slows down severely in some special cases \cite{dir,fei2020iterative}. Correspondingly, this section starts from theoretic computation complexity analysis of IR-NAIC, then a delicate experiment is conducted to verify the assumption that a sufficiently good input context for the language decoder can help reduce the number of iterations. Here we construct the decoder input from the caption hypothesis produced by a well-trained AIC model.

\subsection{Computation Complexity Analysis of IR-NAIC}

For a conventional NAIC model, we assume that the computation cost of each iteration is proportional to the size of input tensor, denoted as $(B \times M, N, H)$, where $B$ is the batch size, $M$ is the beam size, $N$ is the predicted target length, and $H$ is the network dimension. In this way, the total cost of $I$ iteration, generally $I < N$, is $\propto I \times O(B \times M \times N \times H) $. For convenience, we omit $M$ and $H$, which simplify to $I \times O(B \times N)$. Likely, the computation cost for AIC model is $N \times O(B \times 1)$. Note that only decoder with multi-head self-attention needs to consider the previous $i$ words. Then we can define the speedup ratio as:
\begin{equation}
	ratio = \dfrac{N}{I} \times \dfrac{O(B \times 1)}{O(B \times N)}.
\end{equation}
Therefore, fewer iterations and faster parallel computation are the key points to IR-NAIC.

\subsection{Preliminary Experimental Setting}

\paragraph{\textbf{Image Captioning Model. } }
We adopt the official captioning model implementation\footnote{\url{https://github.com/nyu-dl/dl4mt-nonauto}} proposed by \cite{dir} with denoising strategy. For convenience, the regional features of images extracted from faster-rcnn \cite{ren2015faster} on the backone of ResNet-101 \cite{he2016deep} are utilized to retrain the image captioner with the standard Transformer-base configuration \cite{attention}. Our AIC model is first trained with XE loss and then fine-tuned with SCST \cite{Rennie2017Self}.

\paragraph{\textbf{Decoding Strategy. }}
We set the beam size of AIC model to 5 to achieve a good caption hypothesis as input to the decoder. Then we replace a certain percentage of words with \texttt{[mask]} symbol and feed the processed sentence to the iterative refinement decoder. Unlike the standard iterative refinement model that iterates several times, we only iterate fixed one step refinement with the setting of beam size being 1 and substitute all input \texttt{[mask]} symbol with the prediction of the language decoder to determine the final descriptions.

\paragraph{\textbf{Different Masking Methods. }}
In this work, we provide four methods to mask caption hypothesis generated from AIC: $Head$, $Tail$, $Random$ and $Group$. Given the masking rate $p_{mask}$ and the length of caption $N$, the number of masked words can be computed as $N_{mask} = \text{max } (1, N \times p_{mask})$.
Then $Head/Tail$ masking method always masks the first/last $N_{mask}$ words in the sentence, while $Random$ masks the position of caption randomly. $Group$ is slightly different from the above three strategies. It first divides the sentence into $G$ groups, where $G = \text{ceil }(\frac{N}{k})$ and $k$ is the group size. Then in each group, we maintain the first word and mask the remaining $k$-1 words. Thus, the actual masking rate in $Group$ masking can be calculated as:
\begin{equation}
	p_{mask} = 1 - \dfrac{1}{k}.
\end{equation}
To exclude the experimental randomness, we run $Random$ masking method four times with different random seeds and report the average performance result.

\subsection{Results Discussion}

\begin{figure}
	\begin{center}
		\includegraphics[width=1\columnwidth]{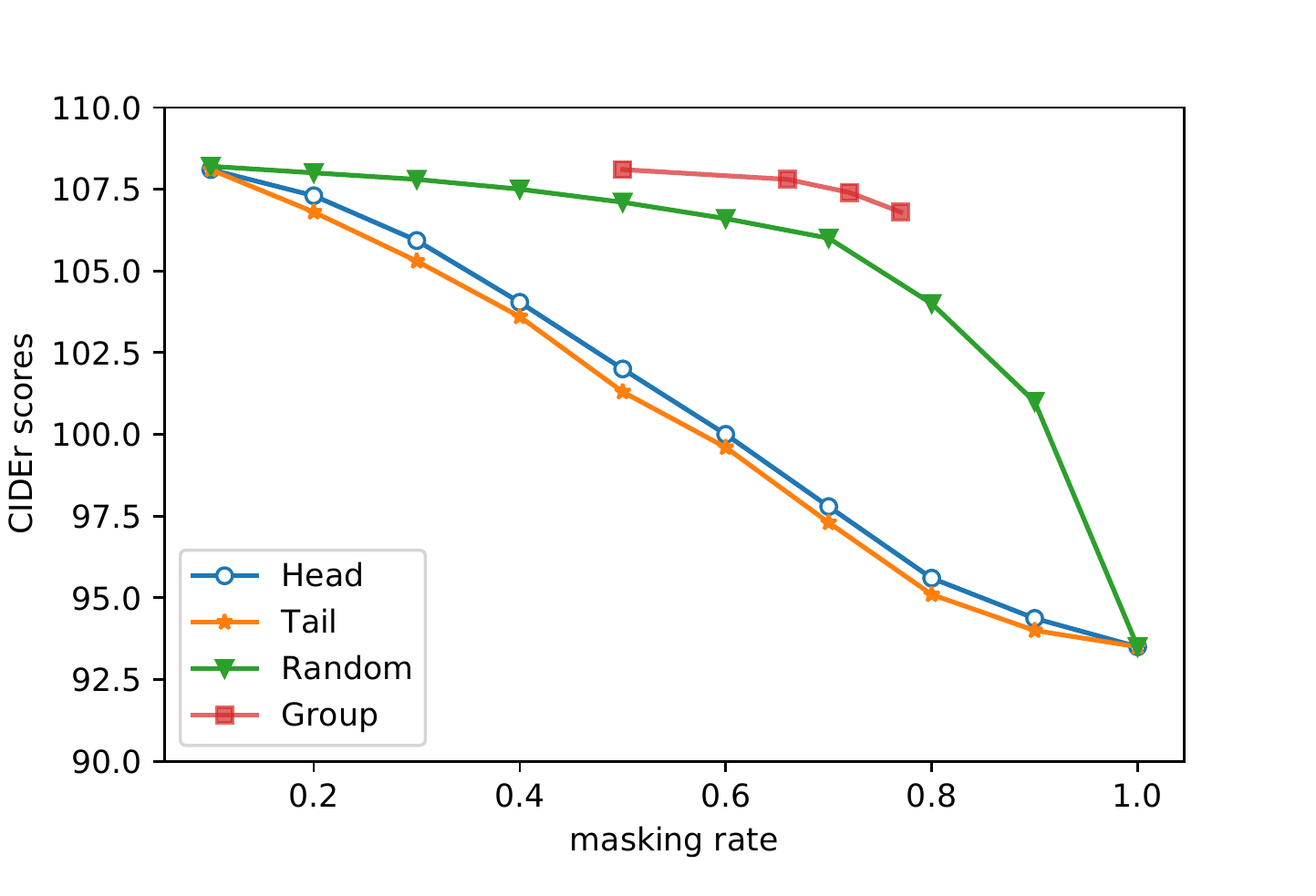}
	\end{center}
	\caption{Performance comparison of four masking methods $\{Head, Tail, Random, Group\}$ under different masking rate in refinement experiments on MS COCO test set.}
	\label{fig:2}
\end{figure}

\begin{table}
	\begin{center}
		\setlength{\tabcolsep}{3.mm}{
		\begin{tabular}{ll|cc}
			\toprule
			$M$&$I$&BLEU-4&CIDEr\\
			\midrule
			\midrule
			1&1&31.0&108.5\\
			5&1&31.2&109.0\\
			1&5&31.0&108.8\\
			5&5&31.3&109.2\\
			\bottomrule
		\end{tabular}}
	\end{center}
	{\caption{Evaluation results of different decoding parameters under $Group$ masking strategy on MS COCO test set. $M$ denotes the beam size, $I$ is the fixed iteration times and $p_{mask}$ is set to 0.7. The metric value does not fluctuate significantly. }
		\label{tab:1}}
\end{table}

The experimental results are shown in Figure \ref{fig:2} and Table \ref{tab:1} respectively, where we can conclude that:

\begin{figure*}
	\begin{center}
		\includegraphics[width=1.9\columnwidth]{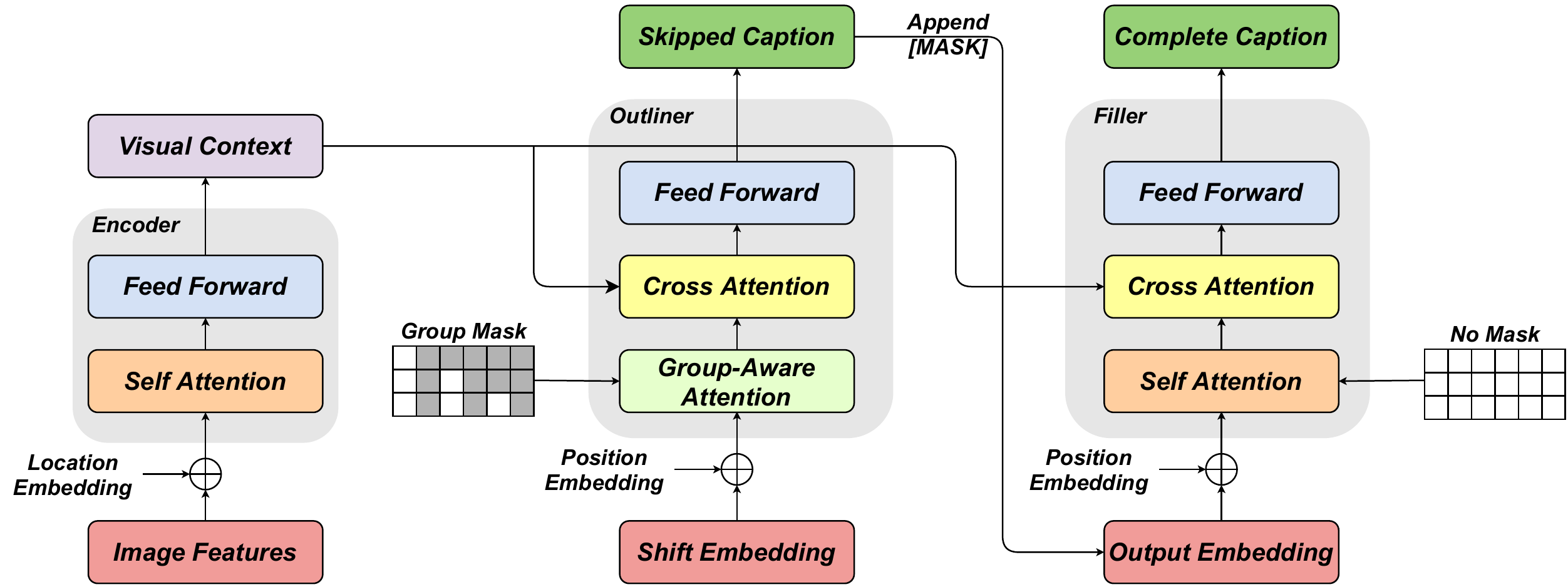}
	\end{center}
	\caption{Overview of the proposed two-stage SAIC framework (group size $k$ = 2), which consists of a visual encoder for image content, an Outliner for skipped caption and a Filler for complete caption. Those two language decoders, sharing the same architecture and parameters, build the sentence in a paradigm where the first word of each group sequentially but remaining words inside in parallel.}
	\label{fig:3}
\end{figure*}

\noindent
\textbf{$\bullet$ Uniform partial prior knowledge is critical. } Compared with $Head$ and $Tail$ masking, it is obvious that both $Random$ and $Group$ can gain a better captioning performance. We attribute it to the fact that the prior knowledge as input to the language decoder is uniformly distributed in $Random$ and $Group$, while the $Head$ and $Tail$ are only provided with concentrated context on one side, \emph{i.e.}, prefix and suffix. $Group$ is superior to $Random$, which indicates that the balanced distribution of deterministic word hypothesis is necessary. $Group$ masking method guarantees that each word without masking can meet at least one deterministic word within the window size of $k$.

\noindent
\textbf{$\bullet$ Small beam size and one iteration are sufficient. } Compared with the standard IR-NAIC model with the beam size of 5 and multiple iterative refinements, it is interesting to find that even if only 30\% of the inputs to the decoder are exposed, the $Group$ masking-based decoder with greedy search can achieve quite comparable performance in fixed 1 iteration.

\section{Approach}

Based on the above analysis, we believe that the iterative refinement in IR-NAIC is unnecessary. In other words, we can obtain a high-quality caption in one shot without going through multi-pass modification under group-aware prior knowledge. In this end, we propose a two-stage framework, referred to as Semi-Autoregressive Image Captioning (SAIC). Briefly speaking, SAIC autoregressively generates a discontinuous sequence with group size $k$ (stage \uppercase\expandafter{\romannumeral1}), and then fills the remaining words (stage \uppercase\expandafter{\romannumeral2}) in one neural network via a non-autoregressive manner. Note that the standard AIC can be regarded as the special case of SAIC when group size equals to 1 and without stage \uppercase\expandafter{\romannumeral2}. 

\subsection{Model Architecture}

\paragraph{\textbf{Overview}} 

Generally, our SAIC model consists of three components, a visual encoder, and two language decoders: Outliner for stage \uppercase\expandafter{\romannumeral1}, and Filler for stage \uppercase\expandafter{\romannumeral2}, as displayed in Figure \ref{fig:3}. All components adopt the Transformer architecture \cite{attention} containing self-attention sublayers and feedforward sublayers. Particularly, additional cross-attention sublayers are added to the two decoders. Furthermore, all of the sublayers are followed by the residual connection and a layer normalization operation.
Note that Outliner and Filler have the same network structure and \emph{share} the parameters, so the number of parameters remains the same as that of standard AIC and NAIC models.

\paragraph{\textbf{Differences from Previous Works}}

Our network architecture takes inspiration from the prior work \cite{dir} and incorporates three key differences with respect to all previous NAIC algorithms as:

(1)  The major difference lies in the \emph{masking strategies} in the self-attention sublayer. The Outliner masks future words discontinuously and causally guarantees a strict left-to-right generation, while the Filler eliminates this limitation to leverage a bi-directional context \cite{devlin2018bert}, as the toy examples of group mask matrix and no mask matrix shown in the Figure \ref{fig:3}.

(2) Our self-attention layer additionally equips with \emph{relative position representation} (RPR) \cite{dai2019transformer} to enable the language decoders to capture the sequence relationship between words easily and effectively. To be specific, the self-attention layer of the language decoders with RPR can be represented by:
\begin{align}
	O_i = \text{softmax}(\dfrac{Q_i(K^T + R_i)}{\sqrt{d_k}})(V + R_i),
	\label{eq:4}\\
	R_i = \text{Emb}(\text{clip}(i, 1), \ldots, \text{clip}(i, N)), \\
	\text{clip}(i,j) = \text{max}(w, \text{min}(w, j-i)),
\end{align}
where $Q$, $K$ and $V$ corresponds to query, key, and value matrices, $d_k$ is the dimension of the key, $R_i$ denotes the relative position embedding matrix, and $w$ denotes the window size, which can also be regarded as the group size $k$.

(3) Most previous NAIC methods need to train the captioning model with an extra independent \emph{length predictor} \cite{fei2019fast,dir,gao2019masked}. However, such a length predictor is implicitly modeled in our Outliner-Filler module because the length of captions is a by-product of autoregressive Outliner, \emph{i.e.}, $N_{fill}=k \times N_{out}$, where $N_{out}$ is the length of sequence produced by Outliner and $N_{fill}$ is the final sequence length from Filler. Another bonus is that we can avoid carefully tuning the weighting coefficient between the length loss and the word prediction loss for performance optimization.

\subsection{Training Strategies}

Directly training our proposed SAIC model is not trivial since a single SAIC model needs to learn to generate captions in both autoregressive and non-autoregressive manners. This section will introduce three training strategies in detail.

\paragraph{\textbf{Group-Aware Sampling}}

Compared with conventional AIC, the length of the caption generated by our Outliner shrinks the caption length from $N_{fill}$ to $N_{out} = \frac{N_{fill}}{k}$. Meanwhile, the masking method in Outliner, referred to as group masking, as illustarted in Figure \ref{fig:3}, is deterministic.  That is, all non-first words are masked in each group. Compared with previous random masking, the training sampling method of our SAIC model is potentially group-aware.

\paragraph{\textbf{Competence-Aware Curriculum Learning}}

Jointly training of Outliner and Filler is problematic since the group-aware training sample cannot make full use of the information about all the words in the sentences. In this work, we propose to gradually transfer from join training \{AIC, NAIC\} to \{Outliner, Filler\} with competence-aware curriculum learning \cite{platanios2019competence}. That is, the captioning model is trained from group size $N_g = 1$ to group size $N_g = k$. More concretely, given a batch of original image-text pairs $(I_i, S_i)_{i=1}^B$ , we first let the proportion of group size $k$ in the Batch be $p_g = 0$ and construct the training samples of AIC and NAIC models for all pairs. And then we gradually increase the group rate $p_g$ to introduce more group-aware learning signals for Outliner and Filler until $p_g =1$. In implementation, we schedule $p_g$ as follows:
\begin{equation}
	p_g = (\dfrac{t}{T})^\lambda, \label{eq:7}
\end{equation}
where $t$ and $T$ are the current and total training steps, respectively. $\lambda$ is a hyperparameter indicating the degree of change and we utilize $\lambda = 1$ to increase the curriculum difficulty $p_g$ linearly.

\paragraph{\textbf{Hybrid Knowledge Distillation}}

In general, the NAIC model usually train the data generated from a teacher AIC model as knowledge distillation due to the smooth data distribution \cite{kim2016sequence,guo2018dynamic}. However, making full use of distillation data may miss the diversity of the raw data. To combine the advantages of both distillation and raw data, we propose a simple yet effective approach -- \emph{hybrid knowledge distillation}. Specifically, hybrid knowledge distillation randomly samples a target sentence from the raw version $S$ with a probability $p_{hybr}$ or its distillation version $S^*$ with a probability $1 - p_{hybr}$ during the entire training stage.

\paragraph{\textbf{Complete Training Algorithm}}

\begin{algorithm}[t] 
	\label{a:1}
	\caption{Semi-Autoregressive Image Captioning Training Algorithm}  
	\KwIn{Training data $D$ including distillation targets, pretrained AIC model $P_{AIC}$, group size $k$, hybrid distillation rate $p_{hybr}$}  
	\KwOut{Semi-Autoregressive Image Captioning model $P_{SAIC}$}  
	$\triangleright$ Fine-tune on pre-trained AIC model\;
	$P_{SAIC}$ $\leftarrow$ $P_{AIC}$\;
	\For{each training batch $(I_i,S_i, S_i^*)_{i=1}^B$}  
	{   $\triangleright$ Hybrid distillation\;
		\For{each sentence $S_i$}
		{Sampling $\hat{S_i} \sim \{S_i, S_i^*\}$ with probability $p_{hybr}$\;}
		$\triangleright$ Curriculum learning\;
		Get the group-aware proportion $p_g$\;
		Split the batch for training into \{AIC, NAIC\} and \{Outliner, Filler\} proportionally\;
		Joint optimization for $P_{SAIC}$\;
	}  
\end{algorithm}

Algorithm 1 describes the procedure of training a SAIC model. To be specific, the SAIC is first initialized by a pre-trained AIC model (Line 2). Then, the training batch randomly selects a raw sentence $S_i$ or its distilled version $S_i^*$ based on probability $p_{hybr}$ (Line 5-7). Next, according to the Equation \ref{eq:7}, we can divide the batch data into two parts: conventional \{AIC, NAIC\} batch $B_{c}$ and group-aware \{Outliner, Filler\} $B_{g}$, where $|\frac{B_{c}}{B_{g}}| = p_g$ (Line 10). We can construct four kinds of training samples based on corresponding data. Finally, we collect all training samples together and accumulate their gradients to update the SAIC model's parameters, which leads to the double of the batch size of standard training.

\begin{table*}[t]
	\begin{center}
			\setlength{\tabcolsep}{3mm}{
		\begin{tabular}{l|cccccc|cc}
			\toprule
			Models&BLEU-1&BLEU-4&METEOR&ROUGE&CIDEr&SPICE&Latency&SpeedUp\\
			\midrule
			\midrule
			\multicolumn{9}{l}{\emph{Autoregressive Image Captioning models}} \\
			\midrule
			NIC-v2 \cite{Vinyals2015Show} &-&32.1&25.7&-&99.8&-&-&-\\
			Up-Down \cite{Anderson2017Bottom}  &79.8&36.3&27.7&56.9&120.1&21.4&-&-\\
			AoANet$^\dagger$ \cite{huang2019attention} &\text{80.2}&\text{38.9}&\text{29.2}&\text{58.8}&\text{129.8}&\text{22.4}&-&-\\ 
			M2-T$^\dagger$\cite{cornia2020meshed} &80.8&39.1&29.2&58.6&131.2&22.6&-&-\\
			AIC$^\dagger$ &80.2&38.7&28.8&58.2&128.7&22.0&185ms&1.00$\times$\\  \midrule
			\multicolumn{9}{l}{\emph{Non-Autoregressive Image Captioning models}} \\  \midrule
			MNIC$^{\dagger \P}$ \cite{gao2019masked} &75.4&30.9&27.5&55.6&108.1&{21.0}&-&2.80$\times$\\ 
			FNIC$^\dagger$ \cite{fei2019fast} &-  &36.2&27.1 & 55.3&115.7&20.2&{\text{-}}&{\text{8.15}}$\times$\\ 
			MIR$^{\dagger \P}$  \cite{dir} &- &32.5&27.2&55.4&109.5&20.6&-&1.56$\times$\\ 
			CMAL$^\dagger$ \cite{guo2020non} &80.3 &37.3&28.1&58.0&124.0&21.8&-&13.90$\times$\\ 
			IBM$^{\dagger \P}$ \cite{fei2020iterative} &77.2 &36.6&27.8&56.2&113.2&20.9&-&3.06$\times$\\  \midrule
			\multicolumn{9}{l}{\emph{Semi-Autoregressive Image Captioning models}} \\  \midrule
			SAIC$^\dagger$ ($M_{out}$= 5, $M_{fill}$= 5) &80.4 &38.7&29.4&58.5&128.3&22.2&119ms&1.55$\times$\\
			SAIC$^\dagger$ ($M_{out}$= 5, $M_{fill}$= 1) &80.4 &38.6&29.3&58.3&127.8&22.1&101ms&1.83$\times$\\
			SAIC$^\dagger$ ($M_{out}$= 1, $M_{fill}$= 1) &80.3 &38.4&29.0&58.1&127.1&21.9&54ms&3.42$\times$\\
			\bottomrule
		\end{tabular}}
	\end{center}
	{\caption{Performance comparisons of different captioning models using different evaluation metrics on the MS COCO Karpathy test set. All values except Latency and SpeedUp are reported as a percentage (\%). “$\dagger$” denotes the model is based on Transformer architecture. AIC is our implementation of autoregressive teacher model, which has the same structure as SAIC. The SpeedUp values of NAIC models are from the corresponding papers. $\P$ represents the NAIC model is iterative refinement-based.}
		\label{tab:3}}
\end{table*}

\subsection{Inference Process}

After encoding the image features, the Outliner starts from \texttt{[bog]} (the beginning of a group) to sequentially generate a jumped subsequence $S_{out} = \{w_1, \ldots, w_m\} $ with group size $k$ until meeting \texttt{[eos]} (end of sentence). Then we construct the input of Filler by appending $k-1$ \texttt{[mask]} symbols after every $w_i$. The final description is generated by replacing all \texttt{[mask]} with the predicted words from Filler with one iteration. If there are multiple \texttt{[eos]} existing, we truncate the sentence where the first \texttt{[eos]} appeared. Note that the beam size $M_{out}$ in Outliner can be different from the beam size $M_{fill}$ in Filler subject to: $M_{out} \geq M_{fill}$. In particular, if $M_{out} > M_{fill} $, then we only feed the Filler with the top $M_{fill}$ hypothesis. Finally, we select the caption hypothesis with the highest score as:
\begin{small}
	\begin{equation}
		\begin{matrix} 
			score(\hat{S})=&\underbrace{\sum_{i=1}^m p(z_i| I, w_{<i})} &+ \underbrace{\sum_{i=0}^{m-1} \sum_{j=1}^{k-1}p(w_{i\times k +j}| I, S_{out})},\\ 
			&\text{Outliner} &\text{Filler} \end{matrix}
	\end{equation}
\end{small}
where $z_i = w_{i \times k}$. Due to the joint training of group size 1 and $k$ simultaneously,  SAIC model can also behave like a standard AIC model by forcing decoding with group size fixed to 1. In this way, we can only use the Outliner to generate the entire sequence without the help of Filler. Thus, AIC $k =1$ can be regarded as a special case.

\begin{table}
	\begin{center}
		\setlength{\tabcolsep}{3mm}{
			\begin{tabular}{lcc}
				\toprule
				Method&Steps&Computing Cost \\
				\midrule 	\midrule
				AIC&$N$&$\sum_{i=1}^{N}Q(i)$\\
				NAIC&1&$Q(1)$\\
				IR-NAIC&$I$&$I \times Q(1)$\\
				SAIC&$\frac{N}{k}+1$&$\sum_{i=1}^{N/k}Q(i) + Q(1)$\\
				\bottomrule
		\end{tabular}}
	\end{center}
	{\caption{Computation complexity analysis for different caption decoding methods. $Q(i)$ denotes the computation cost in autoregressive mode when producing the $i$-th word. Normally, $I$ = 3 $\sim$ 6 and $k$ is set to 3.}
		\label{tab:2}}
\end{table}

\subsection{Computation Complexity Comparison} 

We also provide a theoretic complexity comparison with AIC, one-shot NAIC, IR-NAIC, and our proposed SAIC in Table \ref{tab:2}.
Although both SAIC and AIC models contain a slow generation process, the length of SAIC is $k$ times shorter than conventional AIC. Considering that the computational complexity of self-attention is quadratic with its length, SAIC can save more time during the sequence inference. On the other hand, thanks to the Outliner provides a high-quality semantic context, SAIC does not need to employ a large beam size and multiple iterations like IR-NAIC to maintain a good captioning quality. Experimental results also illustrate that our light Filler can compensate for the extra computation cost in Outliner effectively and can achieve stable acceleration.

\begin{table*}[t]
	\begin{center}
		\setlength{\tabcolsep}{2.5mm}{
		\begin{tabular}{lcccccccccccccc}
			
			\toprule
			&\multicolumn{2}{c}{BLEU-1}&\multicolumn{2}{c}{BLEU-2}&\multicolumn{2}{c}{BLEU-3}&\multicolumn{2}{c}{BLEU-4}&\multicolumn{2}{c}{METEOR}&\multicolumn{2}{c}{ROUGE-L}&\multicolumn{2}{c}{CIDEr}\\ \midrule
			&c5&c40&c5&c40&c5&c40&c5&c40&c5&c40&c5&c40&c5&c40\\
			\midrule
			\midrule
			Up-Down$^*$ \cite{Anderson2017Bottom} &80.2&95.2&64.1&88.8&49.1&79.4&36.9&68.5&27.6&36.7&57.1&72.4&117.9&120.5\\
			AoANet$^*$ \cite{huang2019attention} &81.0&95.0&65.8&89.6&51.4&81.3&39.4&71.2&29.1&38.5&58.9&74.5&126.9&129.6\\
			M2-T$^*$  \cite{cornia2020meshed} &81.6&96.0&66.4&90.8&51.8&82.7&39.7&72.8&29.4&39.0&59.2&74.8&129.3&132.1\\ 
			CMAL  \cite{guo2020non} &79.8&94.3&63.8&87.2&48.8&77.2&36.8&66.1&27.9&36.4&57.6&72.0&119.3&121.2\\ \midrule
			SAIC &80.0&94.5&64.1&88.2&49.2&78.8&37.2&67.8&28.0&36.8&57.7&72.4&121.4&123.7\\ 
			\bottomrule
		\end{tabular}}
	\end{center}
	{\caption{Leaderboard of different image captioning models on the online MS COCO test server. $^*$ denotes the ensemble model.}
		\label{tab:4}}
\end{table*}

\begin{table}
	\begin{center}
		\begin{tabular}{l|ccccc}
			\toprule
			&BLEU-4&METEOR&ROUGE&CIDEr&SPICE\\
			\midrule
			\midrule
			Raw&37.5&28.1&57.3&123.3&21.3\\
			Seq. Dist.&38.4&28.9&58.1&126.6&21.8\\
			Hybr. Dist.&38.4&29.0&58.1&127.1&21.9\\
			\bottomrule
		\end{tabular}
	\end{center}
	{\caption{Performance against different distillation strategies. The results show that our hybrid distillation is superior to the other two distillation strategies. }
		\label{tab:5}}
\end{table}

\begin{table}
	\begin{center}
		\begin{tabular}{c|ccccc}
			\toprule
			$k$&BLEU-4&METEOR&ROUGE&CIDEr&SPICE\\
			\midrule
			\midrule
			2&38.6&29.1&58.3&128.4&22.0\\
			3&38.4&28.9&58.2&127.4&21.9\\
			4&38.4&29.0&58.1&127.1&21.9\\
			\bottomrule
		\end{tabular}
	\end{center}
	{\caption{Evaluation results of different group size $k$. The results show that with the increasing group size $k$, worse quality and faster speed are obtained by our SAIC model.  
			\label{tab:6}}
}\end{table}

\section{Experiments}

\subsection{Experimental Preparation}

\paragraph{\textbf{Dataset}}
MS COCO \cite{chen2015microsoft} is a standard estimation benchmark for image captioning tasks.  To be consistent with previous works, \cite{huang2019attention,cornia2020meshed}, we adopted the Karpathy split \cite{karpathy2015deep} that contains 113,287 training images equipped with 5 human-annotated sentences each and 5,000 images for validation and test splits, respectively.  We omit words that occur less than 5 times. The vocabulary size is 10,369 words. Image features are pre-extracted as \cite{Anderson2017Bottom}. 

\paragraph{\textbf{Evaluation Metrics}}
Six metrics are utilized to comprehensively estimate the model performance: BLEU@$N$ \cite{Papineni2002BLEU}, METEOR \cite{Lavie2007METEOR}, ROUGE-L \cite{Flick2004ROUGE}, CIDEr-D \cite{dis1}, SPICE \cite{spice} and Latency \cite{fei2019fast,guo2020non}. 
Concretely, SPICE focuses on semantic analysis and has a higher correlation with human judgment, and other metrics except Latency favor frequent $n$-grams and measure the overall sentence fluency. Latency is computed based on the time to decode a single sentence without mini batching, and the values are averaged over the whole off-line test set. 

\paragraph{\textbf{Implementation Details}}
Our proposed SAIC model closely follows the same network architecture and hyper-parameters setting as Transformer-base model \cite{attention}. Specifically, the number of stacked blocks is 6, hidden size is 512, and feed-forward Filler size is 2048. 
During training, we first initialize the weights of SAIC model with the pre-trained AIC teacher model. We then train the model for 25 epochs with an initial learning rate of 3e-5 and it decays by 0.9 every 5 epochs. Adam \cite{kingma2014adam} optimizer is employed.  We use the group size $k$ = 4, $\lambda$ = 1 and $p_{hybr} = 0.5$ by default. The decoding time is measured on a single NVIDIA GeForce GTX 1080 TI as prior works reported \cite{guo2020non,fei2020iterative}. 
All speeds are measured by running three times and reporting the average value.

\subsection{Overall Performance}

Table \ref{tab:3} and \ref{tab:4} summarize the performance of both AIC and NAIC models on off-line and on-line MS COCO testing, respectively. According to the evaluation results, we can find that IR-NAIC models, \emph{e.g.}, IBM \cite{fei2020iterative},  outperform those one-pass NAIC models while slow down significantly. However, our small beam variant ($M_{out}$= 1, $M_{fill}$= 1, $k$ = 4) can defeat the existing multiple-iteration models. Also, when the beam size increases to 5, SAIC equipped with a standard Transformer achieves +1.2 CIDEr point improvement compared to the previous best results. We can easily trade-off between performance and speed by using $M_{out}$= 5 and $M_{fill}$= 1. It is somehow surprising that besides NAIC models, SAIC can even outperform the AIC models trained from scratch on some metrics. We attribute it to two reasons: (1) SAIC is fine-tuned on a well-trained AIC, making the training process easier and smoother, (2) Mixing up AIC and NAIC has a better regularization effect than training alone. 
Remarkably, SAIC is consistently achieving good performance with various beam search sizes. These results show that our SAIC is more promising than conventional AIC and IR-NAIC.

\subsection{Model Analysis}

\paragraph{\textbf{Effect of Hybrid Distillation}}

To illustrate the success of incorporating hybrid distillation, we compared different distillation strategies, including raw data (Raw), sequence-level knowledge distillation (Seq. Dist.), and hybrid distillation (Hybr. Dist.) for captioning. The results are listed in Table \ref{tab:5}. Overall, Hybr. Dist. is superior to the other two methods across the board, which indicates that training with raw data and distillation data is complementary. We also find that the performance of distillation data is higher than the raw data, which is consistent with the previous work \cite{guo2020non}.

\paragraph{\textbf{Effect of Group Size}}

We also test the different group size $k$, ranging from \{2,3,4\}, and the results are reported in Table \ref{tab:6}. Obviously, we can find that: (1) A larger $k$ has more significant acceleration on decoding speed because fewer autoregressive steps are required; (2) As $k$ increases, the performance of SAIC drops, \emph{e.g.}, CIDEr with $k = 4$ is 1.3 points lower than that of $k=2$. It illustrates that the learning difficulty of SAIC increases as providing less dependency information.

\subsection{Ablation study}

\begin{table}
	\begin{center}
		\begin{tabular}{c|ccccc}
			\toprule
			&SAIC&-FT&-RPR&-CL&-Hybr. Dist.\\
			\midrule
			\midrule
			BLEU-4&38.4&38.3&38.2&38.0&37.5\\
			CIDEr&127.1&126.8&126.5&126.0&123.3\\
			\bottomrule
		\end{tabular}
	\end{center}
	{\caption{Evaluation results of ablation study. The results show that all introduced techniques help to improve captioning performance effectively.}
		\label{tab:7}}
\end{table}


We also provide an entire ablation study on the MS COCO testing set. As shown in Table \ref{tab:7}, we find that all the proposed help to improve captioning performance to some extent. In particular, the employment of hybrid distillation prevents the SAIC from over-fitting and leads to +3.8 CIDEr score improvement compared to the standard distillation (-Hybr. Dist.). In addition, the other three methods, including training the SAIC model from a pre-trained AIC model (FT), using a relative position representation on decoder (RPR), and using curriculum learning (CL), can bring about 0.3 - 1.1 CIDEr score improvement, respectively.

\subsection{Case Study}
\begin{figure}
	\centering
	\includegraphics[width=1.\columnwidth]{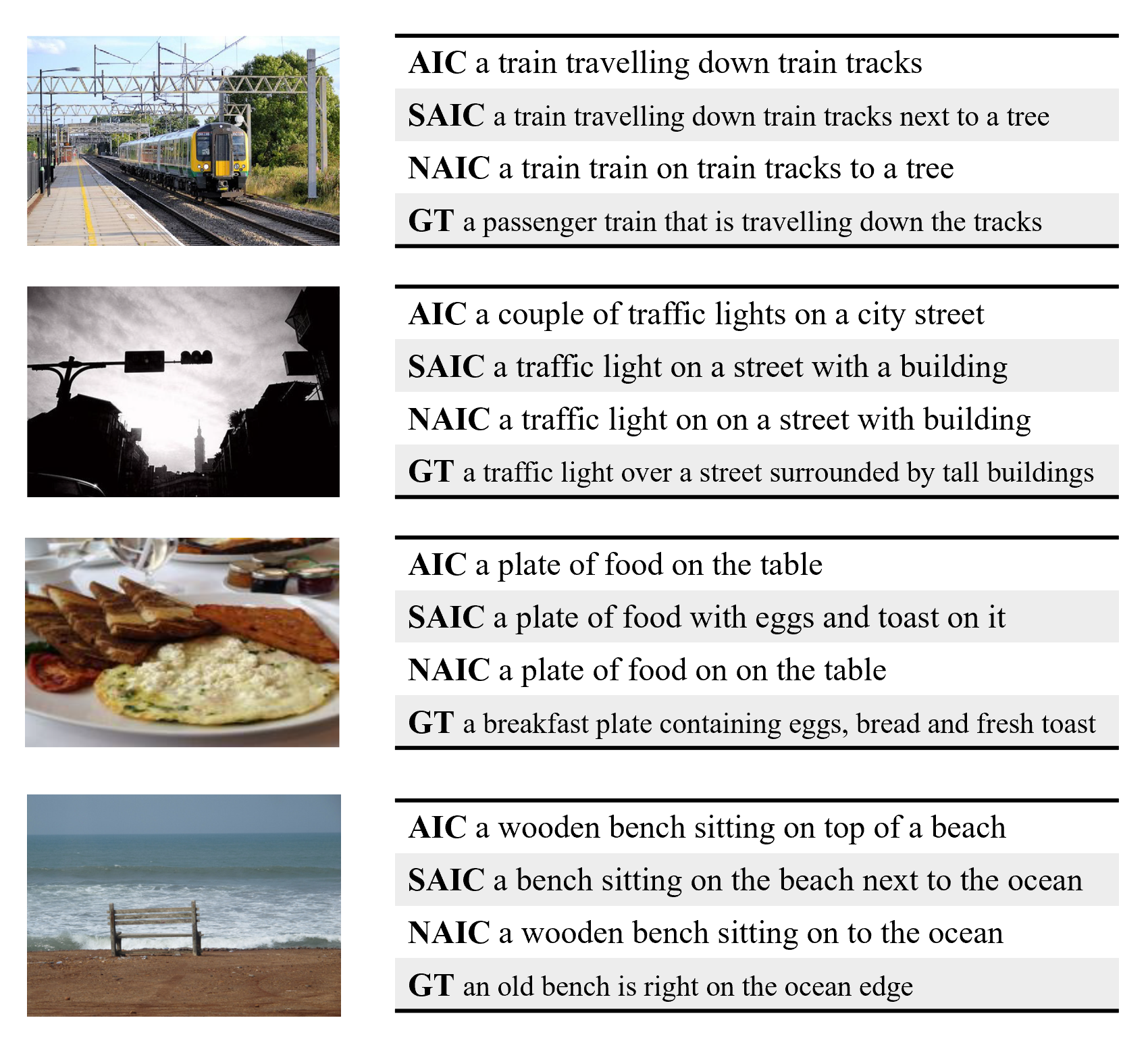}
	\caption{Examples of the generated captions from AIC, SAIC, and NAIC models with the same Transformer architecture. Equipped GT represents human-annotated ground-truth captions. }
	\label{fig:4}
\end{figure}

For more intuitive understanding, we present several examples of generated image captions from AIC, NAIC, and our SAIC ($K$ = 4) models, which hold the same model architecture, coupled with human-annotated ground truth sentences (GT) in Figure \ref{fig:4}. As we can be seen, in general, all captioning models hold the capability to reflect the content of the given image accurately. Meantime,
the incoherent problem, including repeated words and incomplete content, is severe in the sentence generated by pure NAIC, while it can be effectively alleviated by SAIC, \emph{i.e.}, two “train” terms in the first sample. This again confirms that our proposed two-stage framework, including the Outliner and Filler, can guide the captioning model to reduce word prediction errors.

\subsection{Human Evaluation}
Following previous works \cite{huang2019attention,exploring}, we conduct a human study to further compare our SAIC against two patterns, \emph{i.e.}, AIC and NAIC. To be specific, we randomly select 200 samples from the MS COCO testing set and recruit eight workers to compare model performances. Each time, we show only one sentence paired with a corresponding image generated by different approaches or human annotation and ask: can you determine whether the given sentence has been generated by a system or a person? We then calculate the captions that pass the Turing test. The results of Human, SAIC, AIC, and NAIC are 93.6\%, 82.6\%, 81.1\% and 65.0\%, separately. It shows the superiority of SAIC in providing human-like captions.

\section{Related Work}
State-of-the-art image captioning systems are major in autoregressive manner \cite{bai2018survey}, meaning that the model generates captions word by word and is not suitable to modern hardware optimized for parallel execution. The early pioneering work about parallel generation is \cite{zheng2019intention}, which generated the words of selected objects first, and the rest of sentences was filled with a two-pass process. Several recent works attempt to accelerate generation by introducing a NAIC framework \cite{nat,wei2019imitation}, which produces the entire sentences simultaneously. Although accelerating the decoding process significantly, NAIC models suffer from repetitive and missing problems. Therefore, more efforts are devoted to mitigating those issues in the later image captioning work.  
Fei \emph{et al.} \cite{fei2019fast} reorders words detected in the image with a light RNN to form better latent variables before decoding. Cho \emph{et al.}  \cite{dir} and Gao \emph{et al.} \cite{gao2019masked} introduce an iterative mask refinement strategy to learn the position matching information. Lu \emph{et al.} \cite{guo2020non} addresses the inconsistency problem with a multi-agent learning paradigm and sentence-level optimization.
The most relevant to our proposed method is \cite{wang2018semi,hbr} for neural machine translation. 
The biggest difference lies that our model focuses cross-modal information processing and changes one-step generation to a hierarchical form, which maintains a considerable speedup and enables the caption decoder to view more abundant local history and future information to avoid errors. 


\section{Conclusion}

Through a well-designed experiment, we first point out that provided a sufficient decoding prior knowledge, the number of iterations in NAIC could be dramatically reduced. Inspired by this, we propose a two-stage framework, named Semi-Autoregressive Image Captioning, to combine the advantage of AIC and NAIC. In particular, SAIC keeps the sentence outline via an autoregressive manner, and finishes the remaining filling by way of non-autoregression. 
To facilitate model training, we further introduce three strategies, including group-aware sampling, curriculum learning, and hybrid knowledge distillation.
Compared with previously conventional baselines in autoregressive models, our SAIC model achieves a better balance between captioning quality and decoding speed.
Extensive experimental results show that SAIC has an equivalent or even higher performance while maintaining a 50\% faster captioning speed.
For future works, we are curious about how to improve or correct the prior knowledge from the Outliner which correlates well with final outputs. 

\begin{acks}
This work was supported in part by the National Key R\&D Program of China under Grant 2018AAA0102003, in part by National Natural Science Foundation of China: 62022083, 61620106009, 61836002 and 61931008, and in part by Key Research Program of Frontier Sciences, CAS: QYZDJ-SSW-SYS013.
\end{acks}

\bibliographystyle{ACM-Reference-Format}
\balance
\bibliography{cite}
\end{document}